\documentclass[conference]{IEEEtran}
\usepackage{times}

\usepackage[numbers]{natbib}
\usepackage{multicol}

\usepackage{amsmath}
\usepackage{amsfonts}
\usepackage{amssymb}

\usepackage{graphicx}
\usepackage{booktabs}

\usepackage{algorithm}
\usepackage{algpseudocode}

\usepackage{xcolor}

\usepackage{multirow}

\usepackage[bookmarks=true]{hyperref}
\usepackage[nameinlink]{cleveref}
\crefname{section}{Sec.}{Secs.}
\Crefname{section}{Sec.}{Secs.}
\crefname{figure}{Fig.}{Figs.}
\Crefname{figure}{Fig.}{Figs.}
\crefname{table}{Tab.}{Tabs.}
\Crefname{table}{Tab.}{Tabs.}
\crefname{equation}{Eq.}{Eqs.}
\Crefname{equation}{Eq.}{Eqs.}

\usepackage{xspace}
\makeatletter
\DeclareRobustCommand\onedot{\futurelet\@let@token\@onedot}
\def\@onedot{\ifx\@let@token.\else.\null\fi\xspace}

\makeatother

\begin{document}

\title{$\tau_0$-WM: A Unified Video-Action World Model for Robotic Manipulation}




%
\author{
\begin{tabular}{c}
Pengfei Zhou$^{2,*}$ 
Shengcong Chen$^{2,*}$ 
Di Chen$^{2}$ 
Jiaxu Wang$^{2}$ 
Rongjun Jin$^{2}$ 
Bingwen Zhu$^{1,2}$ 
Yike Pan$^{2}$ \\
Songen Gu$^{2}$ 
Kuanning Wang$^{2}$ 
Shufeng Nan$^{2}$ 
Xingyu Qiu$^{2}$ 
Chenhao Qiu$^{2}$ 
Pu Yang$^{2}$ 
Yunuo Cai$^{1,2}$ \\
Jianxiong Gao$^{2}$ 
Yifan Li$^{1}$ 
Yanwei Fu$^{1,2}$ 
Xiangyu Yue$^{2}$ 
Zhi Chen$^{2}$ 
Jianlan Luo$^{1,2\dagger}$ \\
\\[-0.4em]
$^{1}$Shanghai Innovation Institute  $^{2}$AGIBOT Finch \\

\\[-0.4em]
$^{*}$Equal contribution. 
$^{\dagger}$Corresponding author.
\end{tabular}
}

\twocolumn[
  \begin{@twocolumnfalse}
\maketitle
\IEEEpeerreviewmaketitle

    \begin{center}
        \url{https://tau0-wm.github.io/}
    \end{center}

  \end{@twocolumnfalse}
]

\begin{abstract}
Robotic manipulation requires models that generate executable actions while anticipating and evaluating their future consequences before physical execution. We present $\tau_0$-World Model ($\tau_0$-WM), a unified video-action world model that integrates policy learning, video prediction, and action evaluation within a single future-predictive framework. Built on a shared video diffusion backbone, $\tau_0$-WM provides two complementary interfaces. First, a video action model jointly predicts future visual latents and continuous action chunks from multi-view observations, language instructions, and robot state. Second, an action-conditioned video simulator rolls out candidate action chunks into multi-view futures and predicts dense task-progress scores. The model is trained on approximately $27{,}300$ hours of real-robot teleoperation, UMI-style interaction, egocentric human videos, and rollout or failure trajectories using modality-specific supervision masks. At inference time, $\tau_0$-WM uses test-time computation to sample action candidates, rank them with re-denoising consistency, and invoke simulator-based rectification for low-quality candidates. On challenging long-horizon and fine-grained robotic manipulation tasks, $\tau_0$-WM shows superior performance over other relevant baselines. 

\end{abstract}

\section{Introduction}
\label{sec:introduction}

\begin{figure*}
    \centering
    \includegraphics[width=\linewidth]{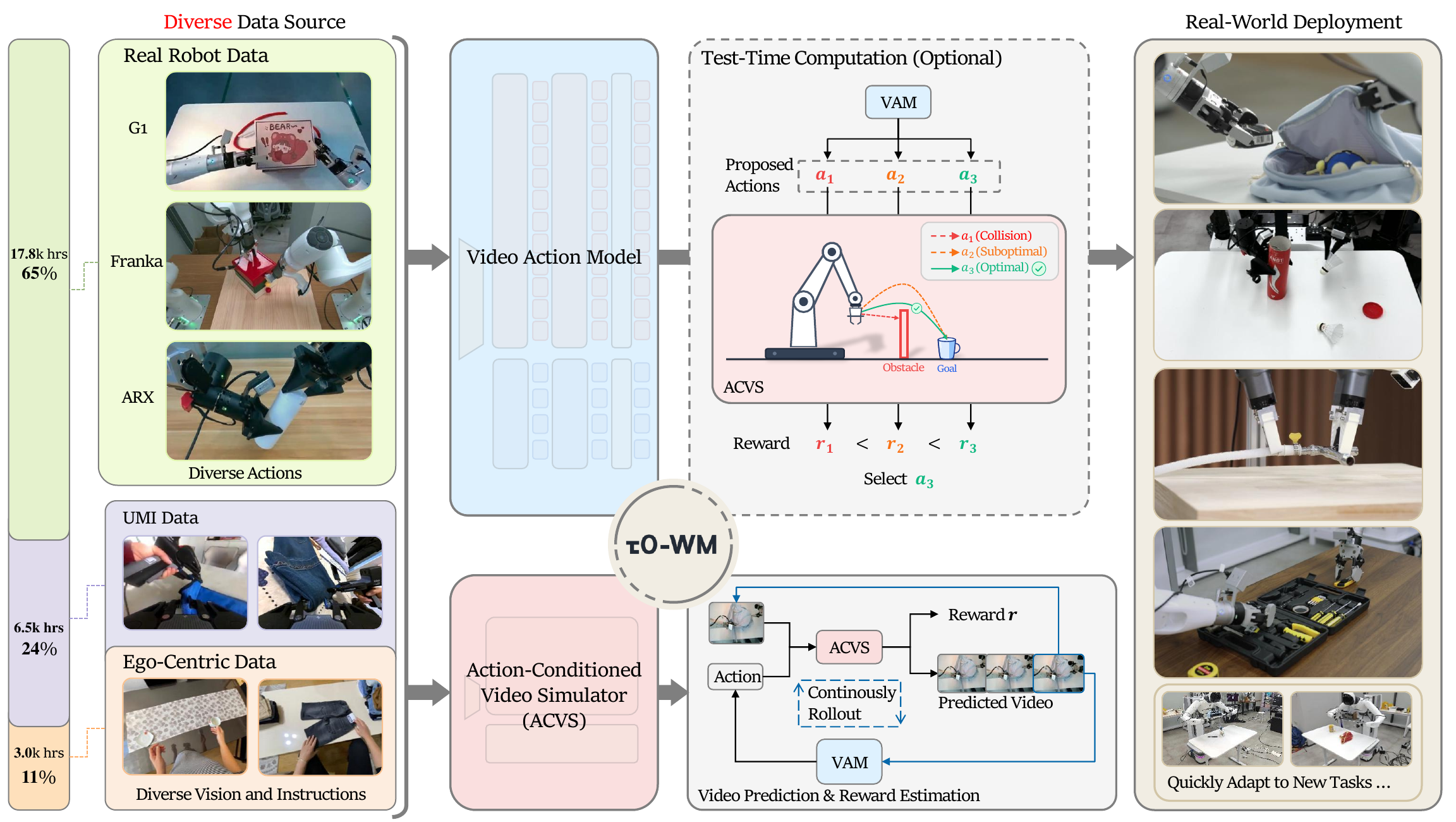}
    \caption{\textbf{Overview of the $\tau_0$-WM framework.} Heterogeneous interaction data from real robots, UMI-style collection, and egocentric human videos are used to train a Video Action Model and an Action-Conditioned Video Simulator. At deployment, the system proposes action candidates, evaluates imagined futures through test-time computation and simulator-based scoring, and selects or rectifies actions for robust manipulation across tasks and embodiments.}
    \label{fig:teaser}
\end{figure*}

Robotic manipulation is fundamentally a problem of acting under uncertain physical consequences. A robot must not only infer which action is likely to satisfy a language instruction, but also anticipate how that action will change the scene through contact, object motion, and multi-step interactions. This perspective connects manipulation policies to a long line of work on predictive control and world models: a useful model should relate observations, actions, and future outcomes in a way that can improve decision making before physical execution~\citep{kalman1960new,hafner2019dreamer,wu2022daydreamer,yang2023unisim,alonso2024diamond}. For real robots, however, this predictive capability must be coupled with an action interface that is executable by a particular embodiment, controller, and sensing stack.

The data needed to learn these two capabilities is available in very different forms. Egocentric videos and human interaction trajectories provide broad evidence about how objects move, how contacts unfold, and how long-horizon tasks are temporally organized. Such data captures rich visual dynamics across diverse objects, scenes, and behaviors, but it does not specify actions in the control space of a deployable robot. Robot demonstrations provide precisely this grounding: they couple observations to continuous actions collected with a specific embodiment, controller, sensor suite, and action representation. Yet robot data is expensive to collect and arguably covers a much narrower subset of objects, environments, tasks. Training only on robot demonstrations yields grounded but narrow policies; training only on broad video data yields predictive but action-ungrounded models. A general manipulation system must therefore use broad interaction data without losing the executable action grounding required for deployment.

This paper studies a unified video-action world modeling formulation for robotic manipulation. The central idea is to place future observations, robot actions, and task progress within a shared predictive model, while allowing each data source to supervise only the signals it actually contains. Video-only data can train visual dynamics; robot trajectories can train executable action generation; progress and failure trajectories can train action-conditioned evaluation. In this way, heterogeneity is not treated as noise or a preprocessing inconvenience, but as a structured source of complementary supervision. The resulting representation is intended to serve not merely as an auxiliary feature for policy learning, but as an interface through which a robot can propose actions, imagine their consequences, and revise them before execution.

We present $\tau_0$-World Model ($\tau_0$-WM), a unified video-action framework that integrates action generation, video prediction, and action-conditioned future evaluation. Rather than separating policy learning from dynamics modeling, $\tau_0$-WM builds both around a shared video diffusion backbone. This backbone exposes two complementary interfaces. The first is a Video Action Model (VAM), which maps multi-view observations, a language instruction, and robot state to both future visual latents and a continuous action chunk. The second is an Action-Conditioned Video Simulator (ACVS), which takes the current observation, instruction, and a candidate action chunk, and predicts the multi-view future rollout together with a dense task-progress trajectory. The distinction between these interfaces is important: VAM answers what the robot should do, while ACVS estimates what would happen if a proposed action were executed.

The shared predictive representation enables $\tau_0$-WM to learn from a heterogeneous corpus of approximately $27{,}300$ hours. This corpus includes real-robot teleoperation data, UMI-style demonstrations, egocentric human videos, and rollout or failure trajectories. These sources provide different degrees of supervision and action fidelity. Real-robot demonstrations provide deployment-aligned continuous actions; UMI-style demonstrations broaden manipulation behaviors and environments with weaker action-like signals; egocentric videos supply large-scale visual interaction dynamics without robot-compatible actions; and rollout or failure trajectories provide supervision for task progress and low-quality outcomes. We train on these sources jointly using modality-specific supervision masks, so that each sample contributes only to the losses supported by its observations, views, states, actions, and progress labels.

At inference time, this unified interface allows $\tau_0$-WM to allocate additional computation to action selection rather than executing the first feed-forward prediction. The model first samples multiple action chunks from VAM and ranks them with a re-denoising consistency score, which measures whether a candidate is consistent with the learned conditional action distribution. When the selected candidate appears unreliable, $\tau_0$-WM invokes ACVS to simulate the futures induced by candidate actions and estimate their task progress. The most promising imagined future is then used to condition a second VAM query, producing a refined action chunk. This yields a proposal--evaluation--revision procedure in which future prediction is used directly as a mechanism for improving robot actions before execution.

We evaluate $\tau_0$-WM as a robot-facing system on fine-grained and long-horizon manipulation tasks across multiple embodiments. In our current evaluation, $\tau_0$-WM achieves the best average success rate among the evaluated baselines on four manipulation tasks. Ablations further show that heterogeneous pre-training improves both zero-shot and fine-tuned performance, while test-time computation improves single-attempt execution through both re-denoising consistency selection and simulator-assisted rectification. These results support the central thesis of this work: video prediction is most useful for robotic manipulation when it is trained jointly with executable action generation and exposed at deployment time as a mechanism for imagining, scoring, and refining future outcomes.

Our contributions are threefold. First, we introduce $\tau_0$-WM, a unified video-action world model that shares a predictive representation across policy learning and action-conditioned simulation. Second, we show how heterogeneous robot, UMI-style, egocentric, and rollout/failure data can be integrated with modality-specific supervision masks. Third, we propose a test-time proposal--evaluation--revision procedure that uses the learned world model to select and rectify actions before execution, improving performance on challenging real-world manipulation tasks.

\section{Related Work}
\label{sec:related_work}

$\tau_0$-WM builds on two related lines of work, robotic video action models and action-conditioned video simulators, and unifies them in a single video-action world-modeling framework. We thus survey works in these two areas and their intersections.

\subsection{Robotic Video Action Models}

Video Action Models (VAMs) introduce future forecasting into robot control by jointly predicting videos and actions~\cite{liao2025genie,bi2025motus,kim2026cosmospolicy,li2026lingbot,yuan2026fast,ye2026gigaworld,li2025unified,zhu2025unified,liang2025video,ye2026world}. Most recent methods build on pretrained video-generation diffusion models~\cite{Wan,hacohen2024ltx,CogVideoX} and adopt a joint-denoising paradigm, where future visual latents and action chunks are generated together~\cite{Cosmos,liao2025genie,bi2025motus,kim2026cosmospolicy,li2026lingbot}. These works show that future prediction provides useful dynamics-aware representations for manipulation. Some recent systems further improve scalability or efficiency, such as Motus~\cite{bi2025motus}, which integrates understanding, video generation, world modeling, and control, and Fast-WAM~\cite{yuan2026fast}, which studies removing future prediction during policy inference to reduce latency.

Different from prior VAMs that mainly use future prediction as an auxiliary policy-learning objective or an optional visual output, $\tau_0$-WM treats video-action modeling as a unified foundation for manipulation. Its VAM jointly predicts multi-view future latents and executable action chunks, while sharing the same predictive representation with an action-conditioned simulator. This enables future prediction to be used not only for representation learning, but also for test-time action evaluation and rectification. Moreover, $\tau_0$-WM is trained on heterogeneous robot, UMI, and egocentric interaction data~\cite{walke2023bridgedata,khazatsky2024droid,chi2024universal,o2024open,hoque2025egodex}, using each data source to supervise the signals it provides.

\subsection{Action-Conditioned Video Simulators for Robotics}
Another line of work uses video models as action-conditioned simulators for decision making. Early visual foresight methods learned action-conditioned video predictors and used model-predictive control to select actions whose predicted futures matched a goal~\cite{finn2017deep,ebert2018visual}. With recent advances in large-scale video generation~\cite{brooks2024sora,veo2024,polyak2024moviegen,runway2025gen4,kong2024hunyuanvideo,Wan}, recent robotics systems condition video models on robot actions, end-effector trajectories, or controllable tokens to predict manipulation rollouts, evaluate policies, or support reinforcement learning~\cite{Cosmos,ali2025world,liao2025genie,guo2025ctrl,jiang2025enerverse,gao2026dreamdojo,chen2022transdreamer}.

In contrast, $\tau_0$-WM does not use the simulator as a separate module. Its Action-Conditioned Video Simulator (ACVS) shares the action interface and backbone configuration with the VAM, is trained on the same heterogeneous data mixture, and predicts both multi-view future rollouts and task-progress scores. At test time, this allows $\tau_0$-WM to go beyond feed-forward action prediction: it samples candidate actions, ranks them by re-denoising consistency, and invokes ACVS to evaluate and rectify low-quality candidates before execution.

\section{Data Sources for Predictive Robot Learning}

A general-purpose Video Action Model should learn not only from a single robot embodiment or data-collection pipeline, but from heterogeneous interaction data that provides complementary forms of supervision. We therefore construct a 27.3K-hour training corpus from three sources: 17.8K hours of real-robot teleoperation on AGIBOT-G01, ARX manipulators, and dual-arm Franka systems; 6.5K hours of filtered open-source UMI-style demonstrations collected with Gen-DAS Grippers~\citep{genrobot2025_realomni}; and 3.0K hours of open-source egocentric human interaction videos~\citep{hoque2025egodex,punamiya2026egoverse,xperience_10m}. These sources differ in embodiment, viewpoint, action fidelity, collection cost, and behavioral diversity, making them naturally suited to different training objectives.

\paragraph{Real-robot teleoperation}

Real-robot demonstrations provide the most reliable action supervision. In our dataset, trajectories are collected on AGIBOT-G01, ARX, and dual-arm Franka platforms across household, retail, and industrial settings, typically with a head-view camera and wrist-mounted cameras. Because these demonstrations are generated directly on robotic systems, their actions are aligned with the robot kinematics, controller interface, sensing stack, and deployment conditions. They are therefore essential for grounding the model in executable robot behavior. At the same time, real-robot data is costly to collect and limited by the available platforms, workspaces, objects, and task setups, which makes it insufficient by itself for broad generalization.

\paragraph{UMI-style demonstrations}

UMI-style data offers a more scalable source of manipulation experience. By using handheld gripper-like devices, human operators can collect demonstrations in diverse environments with substantially lower infrastructure cost than full robot teleoperation. These demonstrations provide rich visual interaction data and action-like signals derived from device motion, which encode useful information about manipulation intent and object interaction. However, these signals are only weakly aligned with deployable robot actions, since the collection device differs from the target robot in embodiment, kinematics, actuation, and control interface. We therefore treat UMI-style demonstrations as scalable but weaker video-action supervision.

\paragraph{Egocentric human interaction videos}

Egocentric human videos provide the broadest coverage of everyday manipulation behaviors. They expose the model to diverse objects, environments, contact patterns, state changes, and long-horizon task structure. Unlike robot or UMI data, however, egocentric videos do not contain robot-compatible action labels and differ substantially in embodiment and viewpoint. Consequently, we use them only for video prediction: they supervise visual dynamics while being excluded from action losses.

\paragraph{Unified supervision}

The three sources induce a hierarchy of supervision. Real-robot data provides deployment-aligned action labels; UMI-style data provides diverse interaction trajectories with weaker action-like signals; and egocentric videos provide large-scale visual dynamics without action supervision. To train on all sources jointly, we use a unified video-action representation with modality-specific supervision masks. For each sample, the mask specifies which inputs are observed, which targets are predicted, and which losses are active. This allows heterogeneous data to contribute to a single end-to-end objective while respecting the different reliability and availability of their supervision.
\section{Video Action Model}
\label{sec:video_action_model}

\begin{figure*}[t]
\centering
\includegraphics[width=\textwidth,trim=0cm 10cm 0cm 4cm,clip]{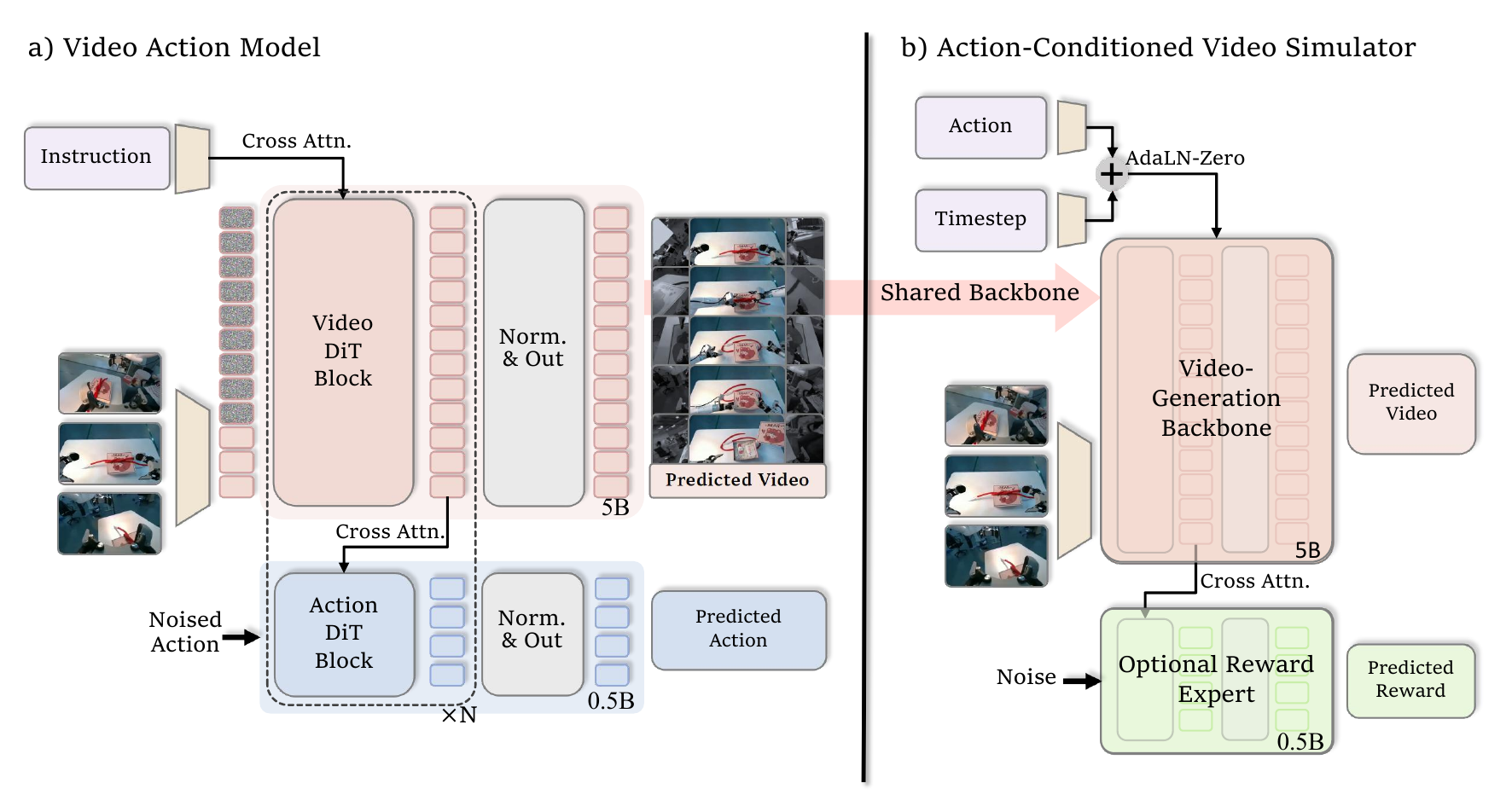}
\caption{\textbf{Architecture of $\tau_0$-WM.}
The Video Action Model (VAM) serves as the policy interface, jointly predicting future visual latents and executable action chunks with a shared video backbone and an Action DiT branch coupled through cross-attention. The Action-Conditioned Video Simulator (ACVS) serves as the evaluation interface, reusing the video-generation backbone to roll out VAM-proposed action chunks and predict dense reward scores for test-time action selection.}
\label{fig:method}
\end{figure*}

\subsection{Model Interface and Problem Formulation}

The Video Action Model (VAM) serves as the policy-facing interface of $\tau_0$-WM. It jointly learns future visual dynamics and executable robot actions using a shared predictive representation. Given the current multi-view observation $\mathbf{o}_t$, language instruction $\mathbf{p}$, and robot state $\mathbf{s}_t$, VAM predicts a future latent trajectory together with an executable action chunk:

\begin{equation}
F_{\theta}(\mathbf{o}_t,\mathbf{p},\mathbf{s}_t)
\rightarrow
\left(
\hat{\mathbf{z}}_{t+1:t+H_v},
\hat{\mathbf{a}}_{t:t+H_a-1}
\right),
\end{equation}
where $\hat{\mathbf{z}}$ denotes future video latents over horizon $H_v$ and $\hat{\mathbf{a}}$ denotes a continuous action chunk over horizon $H_a$. Future visual prediction serves not only as an auxiliary objective but also as a mechanism for learning transferable interaction dynamics from heterogeneous data sources, including videos without action annotations, while action prediction grounds the learned representation in executable robot control.

\subsection{Architecture}
As illustrated in Fig.~\ref{fig:method} (a), VAM consists of two tightly coupled components: a video branch for future visual prediction and an action branch for executable action generation. The two branches share a common predictive representation and interact through feature-level cross-attention, allowing future visual dynamics to directly support action generation.

VAM is instantiated from Wan2.2-TI2V-5B~\cite{Wan}. A Wan VAE first encodes each camera view into latent tensors. For synchronized multi-view inputs, view latents are concatenated along the spatial width dimension, forming a temporally aligned latent canvas. The current observation latent is kept clean as visual context, while future latent slots are noised and denoised by the video branch. The video branch is implemented using the original Wan video DiT backbone (5B parameters) and predicts future latent trajectories through conditional denoising. The action branch is a 0.5B-parameter DiT-style action decoder~\cite{peebles2023scalable} coupled to the video transformer. Together, they form a 5.5B-parameter Video Action Model.

At matched transformer stages, the action tokens first model temporal dependencies within the action horizon and then cross-attend to intermediate video features. These video features are conditioned on both the clean visual context and the language instruction, thereby providing the action branch with instruction-aware and dynamics-relevant visual representations. This feature-level coupling follows recent action-expert designs~\cite{liao2025genie,huang2026enerverse}, while preserving the video backbone as the shared predictive substrate.

\subsection{Joint Flow-Matching Objective}

VAM applies flow matching~\cite{lipman2023flow} to both future video latents and action chunks. Let $\mathbf{z}=\mathbf{z}_{t+1:t+H_v}$ and $\mathbf{a}=\mathbf{a}_{t:t+H_a-1}$ denote the training targets, and let $\mathbf{c}_t$ denote the clean encoded visual context. Given noise levels $u_z$ and $u_a$, the standard flow-matching construction produces noised inputs $\tilde{\mathbf{z}}, \tilde{\mathbf{a}}$ together with velocity targets $\mathbf{v}_{\mathbf{z}}, \mathbf{v}_{\mathbf{a}}$. We optimize
 
\begin{equation}
\begin{aligned}
\mathcal{L}_{\mathrm{VAM}}
= \mathbb{E}\Big[
&\lambda_z
\left\|
f_{\theta}^{z}(\tilde{\mathbf{z}},u_z,\mathbf{c}_t,\mathbf{p})
-\mathbf{v}_{\mathbf{z}}
\right\|_2^2  \\
&+
\lambda_a
\left\|
f_{\theta}^{a}(\tilde{\mathbf{a}},u_a,\mathbf{s}_t,\mathbf{h})
-\mathbf{v}_{\mathbf{a}}
\right\|_2^2
\Big],
\end{aligned}
\end{equation}
where $f_{\theta}^{z}$ and $f_{\theta}^{a}$ denote the video and action vector-field heads, and $\mathbf{h}$ denotes the intermediate video features consumed by the action branch.

The expectation is taken over heterogeneous training samples with different supervision levels. Robot trajectories contribute both visual prediction and action supervision, while egocentric human videos contribute only the visual dynamics term. Missing modalities are handled through supervision masks, allowing all data sources to participate in a unified training process. In all experiments, we simply set $\lambda_z=\lambda_a=1$.

\subsection{Inference and Deployment}

At inference time, VAM takes the latest multi-view observation $\mathbf{o}_t$, language instruction $\mathbf{p}$, and robot state $\mathbf{s}_t$ as input and predicts an executable action chunk. The future latents can be decoded into video frames when explicit visual rollouts are required, or retained as latent representations when used solely to support action generation. This enables two deployment modes. In action-only deployment, only the predicted action chunk is generated and executed in a receding-horizon manner, providing efficient real-time control. In rollout-enabled deployment, VAM additionally predicts future visual latents that can be decoded into multi-view videos, allowing future scene evolution to be explicitly visualized when desired.

\section{Action-Conditioned Video Simulator}
\label{sec:video_simulator}

\subsection{Simulator Interface and Problem Formulation}

The Action-Conditioned Video Simulator (ACVS) serves as the evaluation interface of $\tau_0$-WM, as illustrated in Fig.~\ref{fig:method} (b). Whereas VAM proposes executable action chunks, ACVS estimates the future consequences induced by a candidate action. Instead of physically executing every candidate action on the robot, ACVS predicts future visual rollouts and dense reward trajectories, providing an action-conditioned proxy for deployment-time evaluation.

Given memory observations $\mathbf{o}_{t-M:t}$, a language instruction $\mathbf{p}$, and a candidate action chunk $\bar{\mathbf{a}}_{t:t+H_a-1}$, ACVS predicts future video latents together with dense reward scores:

\begin{equation}
G_{\phi}
(
\mathbf{o}_{t-M:t},
\mathbf{p},
\bar{\mathbf{a}}_{t:t+H_a-1}
)
\rightarrow
\left(
\hat{\mathbf{z}}_{t+1:t+H_v},
\hat{\mathbf{r}}_{t:t+H_a-1}
\right),
\end{equation}
where $\hat{\mathbf{z}}$ denotes the imagined future latent rollout and $\hat{\mathbf{r}}$ denotes the predicted reward trajectory. ACVS is not an action policy; it treats the candidate action chunk as a clean condition and evaluates the future it induces.

\subsection{Architecture}

ACVS reuses the Wan VAE and video transformer backbone~\cite{Wan} but removes the Action DiT policy branch. Memory and current observations are encoded into clean latent context, while future latent slots are initialized with noise and denoised by the video backbone.

To condition future prediction on candidate actions, we follow the action-conditioned design of Cosmos~\cite{ali2025world}. For each future latent slot $\ell$, temporally aligned actions are grouped into an action block $\mathbf{b}_{\ell}$ and projected through lightweight MLPs:

\begin{equation}
\mathbf{c}^{a}_{\ell}
=
\psi_D(\mathbf{b}_{\ell}),
\qquad
\mathbf{m}^{a}_{\ell}
=
\psi_{6D}(\mathbf{b}_{\ell}),
\end{equation}
which are injected into the diffusion-time embedding and AdaLN modulation embedding, respectively. The resulting action conditions are broadcast across spatial tokens and camera views for the corresponding future slot, while observation slots remain unconditioned.

Unlike VAM, ACVS does not generate actions. Its sole purpose is to estimate how the scene would evolve under a proposed action sequence, allowing different candidate actions to induce different imagined futures under the same observation and instruction.

\subsection{Reward and Progress Scoring}

In addition to predicting future visual rollouts, ACVS predicts a dense reward trajectory for each candidate action chunk. We decompose each manipulation task into subtasks and assign progress labels at the subtask level. Frame-level rewards are then estimated through Monte Carlo propagation within each subtask segment, producing dense supervision rather than a single terminal success label.

Failure data is intentionally incorporated into reward construction. For failed subtask segments, the reward is assigned a negative value across the corresponding trajectory. These failure examples teach ACVS to identify action-conditioned futures that lead to unsuccessful contact, incorrect object motion, or task regression. Consequently, ACVS learns to distinguish actions that make meaningful task progress from those that merely produce visually plausible motion.

To further improve simulator fidelity, we augment simulator training with failure-heavy and recovery trajectories. While such data may be suboptimal as direct policy supervision, it is particularly valuable for simulator learning because it exposes the model to off-distribution actions, failed interactions, and recovery behaviors that are difficult to observe from successful demonstrations alone.

\subsection{Training Objective}

ACVS uses the same flow-matching formulation as VAM and jointly supervises future video latents and dense reward trajectories. Let $\mathbf{c}_{t-M:t}$ denote the clean visual context, $\mathbf{z}_{t+1:t+H_v}$ the future latent rollout induced by candidate action $\bar{\mathbf{a}}$, and $\mathbf{r}_{t:t+H_a-1}$ the target reward trajectory. Given noise levels $u_z$ and $u_r$, the standard flow-matching construction produces noised inputs $\tilde{\mathbf{z}},\tilde{\mathbf{r}}$ together with velocity targets $\mathbf{v}_{\mathbf{z}},\mathbf{v}_{\mathbf{r}}$. We optimize

\begin{equation}
\begin{aligned}
\mathcal{L}_{\mathrm{ACVS}}
=
\mathbb{E}\Big[
&\lambda_z
\left\|
g_{\phi}^{z}
(\tilde{\mathbf{z}},u_z,\mathbf{c}_{t-M:t},\mathbf{p},\bar{\mathbf{a}})
-\mathbf{v}_{\mathbf{z}}
\right\|_2^2
\\
&+
\lambda_r
\left\|
g_{\phi}^{r}
(\tilde{\mathbf{r}},u_r,\mathbf{h})
-\mathbf{v}_{\mathbf{r}}
\right\|_2^2
\Big],
\end{aligned}
\end{equation}
where $g_{\phi}^{z}$ and $g_{\phi}^{r}$ denote the video and reward velocity predictors, respectively, and $\mathbf{h}$ denotes the action-conditioned video features consumed by the reward head. In all experiments, we simply set $\lambda_z=\lambda_r=1$.

\section{Test-Time Computation}
\label{sec:test_time_computation}

Pre-training on large-scale heterogeneous interaction data makes the conditional action distribution inherently multimodal: for the same instruction and scene, the robot may complete the task through multiple feasible action sequences. These solutions can differ in precision, robustness, and likelihood of success. Consequently, selecting a high-quality action becomes an important deployment-time problem.

To address this challenge, $\tau_0$-WM adopts a coarse-to-fine test-time computation strategy. It first samples multiple action candidates from VAM and applies a lightweight self-consistency filter to identify reliable candidates. Only when the sampled candidates appear unreliable does the system invoke ACVS for more expensive rollout-based evaluation and action rectification. This design preserves real-time performance in most situations while retaining the ability to recover from difficult states. The overall procedure is summarized in Alg.~\ref{alg:ttc}.

\begin{algorithm}[t]
\caption{Test-Time Computation}
\label{alg:ttc}
\begin{algorithmic}[1]

\Require VAM $F_\theta$, ACVS $G_\phi$
\Require current context $\mathcal{C}_t$
\Require candidate budget $N$, threshold $\gamma$

\State Sample $N$ candidate actions
$\{\bar{\mathbf{a}}^{(i)}\}_{i=1}^{N}$

\For{$i=1,\dots,N$}
    \State Compute
    $S_{\mathrm{RCS}}^{(i)}$
\EndFor

\State
$i^\star
\leftarrow
\arg\max_i S_{\mathrm{RCS}}^{(i)}$

\If{$S_{\mathrm{RCS}}^{(i^\star)} \ge \gamma$}
    \State \Return $\bar{\mathbf{a}}^{(i^\star)}$
\EndIf

\For{$i=1,\dots,N$}
    \State Evaluate candidate using ACVS
    \State Compute rollout value $J^{(i)}$
\EndFor

\State
$j^\star
\leftarrow
\arg\max_i J^{(i)}$

\State \Return
$\mathrm{LAR}
(\hat{\mathbf{z}}^{(j^\star)})$

\end{algorithmic}
\end{algorithm}

\begin{algorithm}[t]
\caption{Low-quality Action Rectification}
\label{alg:lar}
\begin{algorithmic}[1]

\Require selected rollout latent
$\hat{\mathbf{z}}^{(j^\star)}$

\State Convert
$\hat{\mathbf{z}}^{(j^\star)}$
into future conditioning

\State Re-query VAM with
$\mathbf{o}_t,\mathbf{p},\mathbf{s}_t$
and the selected future condition

\State Generate refined action chunk
$\tilde{\mathbf{a}}$

\State \Return $\tilde{\mathbf{a}}$

\end{algorithmic}
\end{algorithm}

\subsection{Re-denoising Consistency Score}

Given the current context
$\mathcal{C}_t=(\mathbf{o}_t,\mathbf{p},\mathbf{s}_t)$,
VAM samples $N$ candidate action chunks
$\{\bar{\mathbf{a}}^{(i)}\}_{i=1}^{N}$.

For each candidate, we randomly sample $K$ flow timesteps and re-noise the action according to the same flow-matching process used during training. The re-noised action is then evaluated by VAM's action vector field, producing an average re-denoising error $\mathcal{E}_{\mathrm{RCS}}^{(i)}$.

We define the Re-denoising Consistency Score (RCS) as

\begin{equation}
S_{\mathrm{RCS}}^{(i)}
=
-\mathcal{E}_{\mathrm{RCS}}^{(i)},
\end{equation}
and select

\begin{equation}
i^{\star}
=
\arg\max_i
S_{\mathrm{RCS}}^{(i)}.
\end{equation}

RCS serves as a lightweight distributional filter. It favors candidates that are more consistent with the learned conditional action manifold while introducing negligible computational overhead compared with rollout-based evaluation.

\subsection{Low-quality Action Rectification}

Although RCS identifies the most self-consistent candidate among the sampled actions, all candidates may still be poor in challenging states. We therefore introduce Low-quality Action Rectification (LAR).

When the selected candidate satisfies

\begin{equation}
S_{\mathrm{RCS}}^{(i^\star)} < \gamma,
\end{equation}
where $\gamma$ denotes a reliability threshold, ACVS is invoked to evaluate all candidate actions. For each candidate action chunk, ACVS predicts an imagined rollout and a dense reward trajectory

\begin{equation}
(\hat{\mathbf{z}}^{(i)},\hat{\mathbf{r}}^{(i)})
=
G_{\phi}
(
\mathbf{o}_{t-M:t},
\mathbf{p},
\bar{\mathbf{a}}^{(i)}
).
\end{equation}

The rollout value is computed as

\begin{equation}
J^{(i)}
=
\max_{0 \le q < H_a}
\hat{r}^{(i)}_{t+q},
\end{equation}
where $J^{(i)}$ measures the maximum task progress achieved by the imagined rollout. The highest-value rollout

\begin{equation}
j^\star
=
\arg\max_i J^{(i)}
\end{equation}
is selected as the most promising future.

\begin{figure*}[!t]
    \centering
    \vspace*{-0.5em}
    \includegraphics[
        width=\textwidth,
        keepaspectratio
    ]{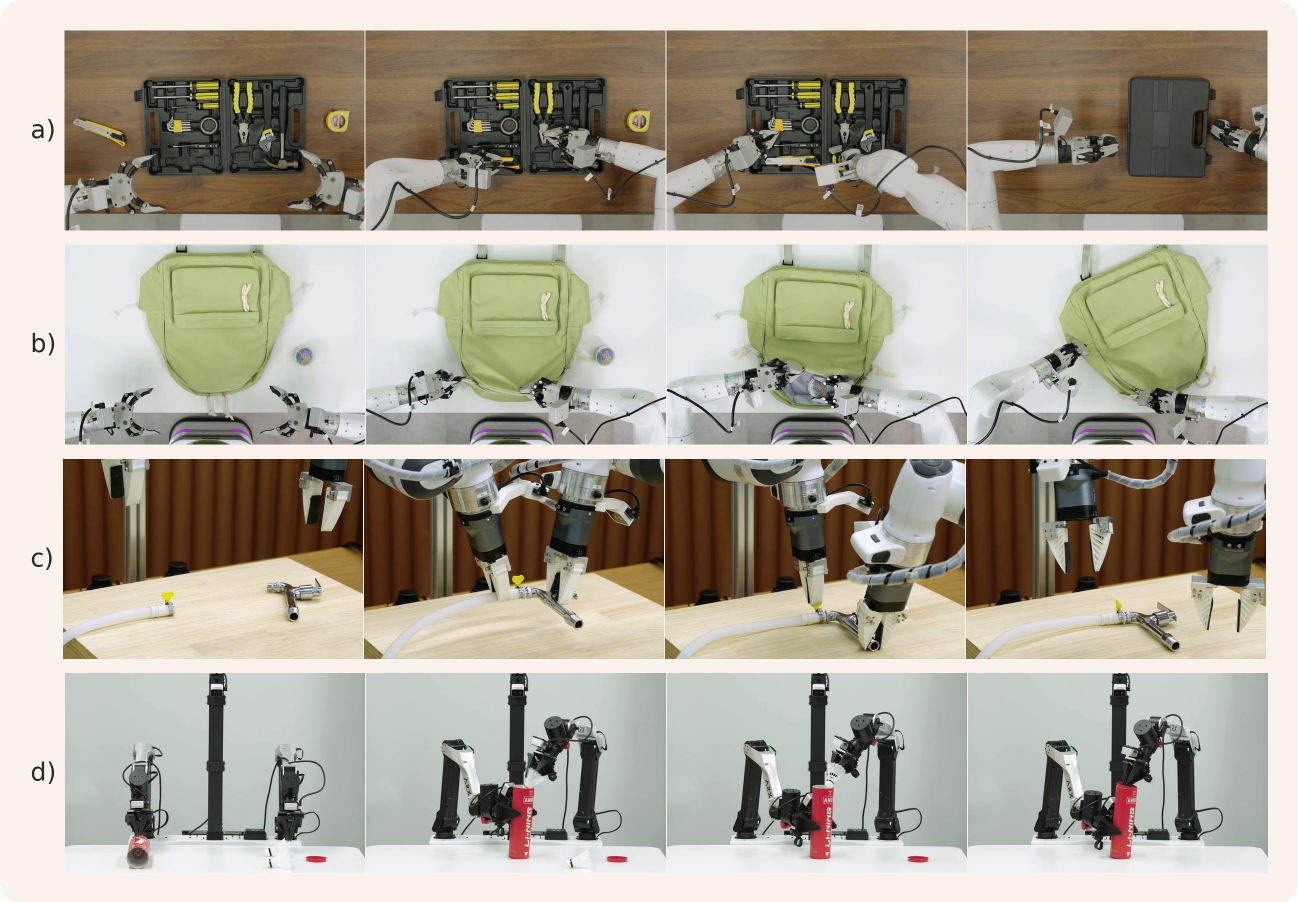}
    \vspace{-0.5em}
    \caption{\textbf{Illustrations of our evaluation tasks.}
    (a) Storing different tools on the desk into their corresponding places in the toolbox (\textit{Toolbox}). 
    (b) Unzipping the school bag, storing objects into it, and zipping up (\textit{School Bag}). 
    (c) Connecting the hose to the faucet and securing it (\textit{Faucet}).
    (d) Storing the badminton shuttlecocks and closing the lid (\textit{Badminton}).}
    \label{fig:filmstream}
\end{figure*}

Instead of directly executing the corresponding action, we perform a second policy query conditioned on the selected future rollout. Specifically, the rollout latent $\hat{\mathbf{z}}^{(j^\star)}$ is converted into an additional future condition and injected into VAM, allowing the policy to generate a refined action chunk that is explicitly guided toward the selected high-value future. The rectification procedure is summarized in Alg.~\ref{alg:lar}.

\section{Experimental Evaluation}
\label{sec:experiments}

We evaluate $\tau_0$-WM on long-horizon, fine-grained real-robot manipulation tasks. Our experiments aim to answer three questions: (i) whether the proposed VAM enables strong policy performance on challenging multi-stage manipulation tasks, (ii) whether heterogeneous pre-training with robot, UMI-style, and egocentric interaction data improves downstream performance, and (iii) whether deployment-time computation further improves closed-loop execution through action selection and rectification. We compare against representative policy and video-action baselines and conduct ablations on both data composition and test-time computation.

\noindent \textbf{Experimental setup.}
Experiments span three robot embodiments---AGIBOT-G01, ARX manipulators, and a dual-arm Franka system---and include language-conditioned, multi-view packing and assembly tasks. The primary evaluation metric is task success rate. We compare $\tau_0$-WM against representative policy and video-action baselines, including $\pi_{0.5}$~\cite{black2025pi05} and Fast-WAM~\cite{yuan2026fast}. For deployment-time reasoning, we additionally compare our test-time computation strategy against standard execution, classifier-free guidance (CFG)~\cite{ho2022classifier}, and Action Coherence Guidance (ACG)~\cite{park2025acg}. Additional implementation details, including training hyperparameters, deployment latency, and inference settings, are provided in the appendix.

\subsection{Main Results}

\begin{figure*}[t]
    \centering
    \includegraphics[width=\textwidth]{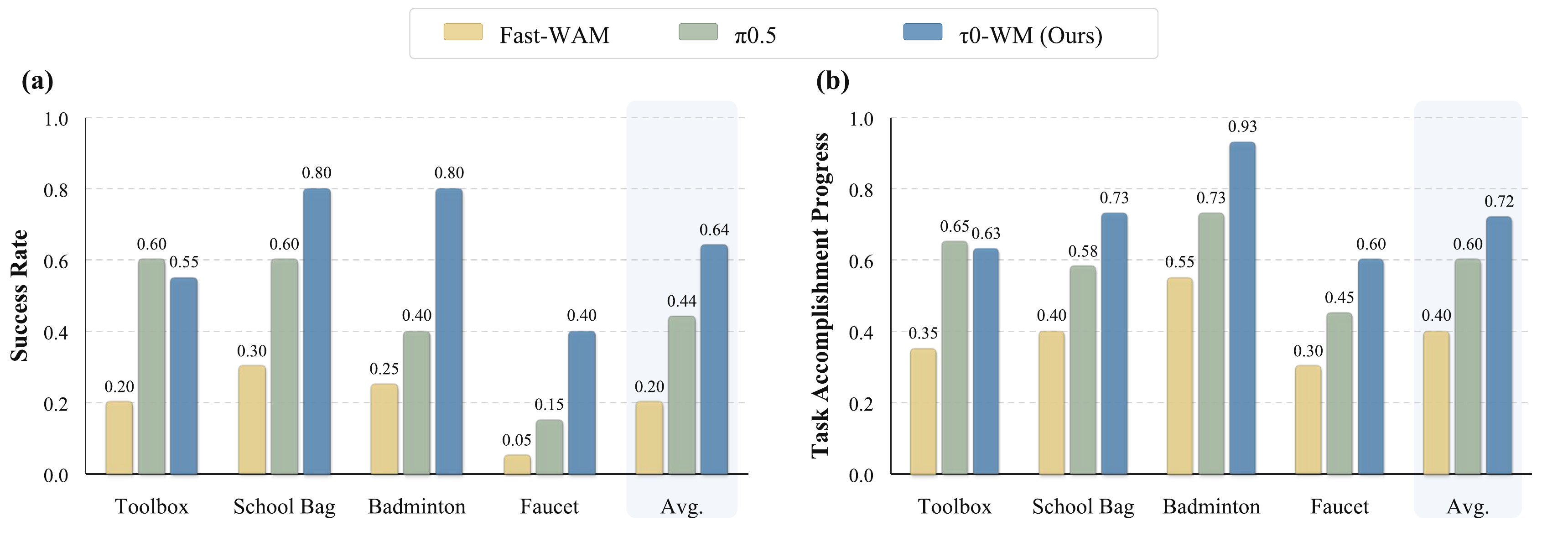}
    \caption{\textbf{Comparison of different models in terms of success rate and task accomplishment progress.}
    Considering the complexity of the long-horizon tasks, we evaluate different models using both task success rate and stepwise task accomplishment progress.}
    \label{fig:mainres}
\end{figure*}

\begin{table}[!t]
    \centering
    \caption{\textbf{Effect of Ego and UMI pre-training.}
    Success rates on zero-shot and SFT evaluation for different pretraining recipe.}
    \label{tab:umi-ego-ablation}
    \begin{tabular}{lccc}
        \toprule
        \multicolumn{4}{c}{\textbf{Zero-shot: Pen-to-holder}} \\
        \midrule
        Data & Clean & Clut. & Avg. \\
        \midrule
        Robot & 0.22 & 0.06 & 0.14 \\
        Robot+UMI+Ego & \textbf{0.56} & \textbf{0.53} & \textbf{0.55} \\
        \midrule
        \midrule
        \multicolumn{4}{c}{\textbf{SFT: Object-wipe-place}} \\
        \midrule
        Data & Clean & Clut. & Avg. \\
        \midrule
        Robot & 0.85 & 0.55 & 0.70 \\
        Robot+UMI+Ego & \textbf{0.90} & \textbf{0.75} & \textbf{0.83} \\
        \bottomrule
    \end{tabular}
\end{table}

We evaluate closed-loop execution on four precision-sensitive manipulation tasks shown in Fig.~\ref{fig:filmstream}, all of which are excluded from the pre-training corpus. These tasks require long-horizon reasoning, multi-stage object interaction, and precise geometric alignment. For example, \textit{School Bag} requires sequential zipper manipulation and object placement, while \textit{Faucet} requires accurate hose alignment and secure attachment. To evaluate embodiment diversity, \textit{Badminton} is conducted on the ARX manipulator and \textit{Faucet} on the dual-arm Franka platform, while the remaining tasks are performed on AGIBOT-G01.

Fig.~\ref{fig:mainres} shows that $\tau_0$-WM achieves the highest average success rate and performs best on most tasks. Although $\pi_{0.5}$ performs competitively on \textit{Toolbox}, its performance degrades on tasks requiring longer-horizon coordination and fine-grained manipulation. In contrast, $\tau_0$-WM remains consistently strong across all four tasks. Notably, \textit{Faucet} remains challenging for every method, indicating that the task is far from saturated; nevertheless, $\tau_0$-WM achieves the highest success rate under these strict alignment constraints. We attribute this advantage to the joint modeling of future visual dynamics and executable actions, which provides richer predictive supervision than action-only policy learning. Overall, these results demonstrate that VAM scales effectively across multiple robot embodiments while maintaining strong long-horizon manipulation capability.

Interestingly, we also observe qualitative differences that are not captured by the binary success metric. In the \textit{Toolbox} task, baseline policies frequently stop once a tool is inserted into the correct slot, even when the insertion is incomplete or the tool remains loosely positioned. By contrast, $\tau_0$-WM often performs additional corrective actions, such as pushing or pressing the tool further into place, before terminating the episode. We hypothesize that this behavior emerges from the explicit modeling of future visual outcomes, which encourages the policy to optimize for the quality of the final scene configuration rather than merely reaching an intermediate task-completion state.


\begin{table}[!t]
    \centering
    \caption{\textbf{Comparison between Test-time Computation Variants.} Success rate is reported for each task and averaged across tasks. One-time completion only. Retries are not allowed.}
    \label{tab:test-time-compute}
    \begin{tabular}{l|ccc}
        \toprule
        Variant & Tissue$\rightarrow$Box & Pen$\rightarrow$Box & Avg. \\
        \midrule
        w/o TTC & 0.55 & 0.30 & 0.43 \\
        w. CFG~\cite{ho2022classifier} & 0.25 & 0.15 & 0.20 \\
        w. ACG~\cite{park2025acg} & 0.40 & 0.35 & 0.38 \\
        \midrule
        w. RCS & 0.65 & 0.35 & 0.50 \\
        w. RCS + LAR & \textbf{0.70} & \textbf{0.50} & \textbf{0.60} \\
        \bottomrule
    \end{tabular}
\end{table}

\subsection{Ablation Studies}

\noindent \textbf{Pre-training data composition.} To validate the effectiveness of the proposed pretraining data mixture, we trained two $\tau_0$-WM models: a model trained exclusively on robot teleoperation data, and a model trained on the complete pretraining corpus.
We conduct the comparison under both zero-shot execution and supervised fine-tuning. The zero-shot task requires picking up a pen and placing it into a pen holder, while the fine-tuning task requires picking up an object, wiping off dirt, and returning it to the tabletop. Both tasks are evaluated in clean and cluttered tabletop variants.

Table~\ref{tab:umi-ego-ablation} shows that adding UMI and egocentric data improves performance in both settings. The gain is most pronounced in the zero-shot setting, where success rate improves from 0.14 to 0.55 on average. This suggests that UMI and egocentric interaction data primarily improve general-purpose manipulation priors and visual understanding, which transfer effectively to previously unseen tasks. The benefit remains visible after SFT, particularly under cluttered conditions, indicating improved robustness rather than merely faster adaptation.

\noindent \textbf{Test-time computation.}
We ablate the proposed test-time computation strategy on two tasks based on the pretrained VAM: pulling out a tissue and placing it into a box (\textit{Tissue$\rightarrow$Box}) and picking up a pen and placing it into a box (\textit{Pen$\rightarrow$Box}). To isolate the effect of TTC, we adopt a stricter protocol that allows only a single attempt without retries. Each experiment is repeated 20 times.

Unless otherwise specified, TTC uses four action proposals per decision step. It consists of two stages: \textit{RCS}, which performs lightweight self-consistency-based candidate selection, and \textit{LAR}, which invokes ACVS for rollout-based action rectification when the selected action is deemed unreliable.

Table~\ref{tab:test-time-compute} shows that both stages of the proposed test-time computation improve execution performance under the single-attempt setting. Using only the lightweight RCS filter increases the average success rate from 0.43 to 0.50, indicating that a substantial portion of failures originates from selecting suboptimal action samples rather than insufficient policy capability. Further enabling LAR improves the average success rate to 0.60 by leveraging action-conditioned future rollouts for action rectification.

Table~\ref{tab:test-time-compute} also compares our method with existing generation-time guidance approaches. While CFG~\cite{ho2022classifier} and ACG~\cite{park2025acg} modify the generation process itself, our approach explicitly evaluates candidate actions and their induced futures before execution. As a result, RCS+LAR consistently achieves the best performance across all tasks. The larger improvement on \textit{Pen$\rightarrow$Box} further suggests that future-conditioned rectification is particularly beneficial for manipulation tasks that require precise object placement and alignment.


\section{Conclusion and Future Work}
\label{sec:conclusion}

We presented $\tau_0$-WM, a unified video-action world model that combines action generation, future prediction, and deployment-time reasoning within a single predictive framework.

A key aspect of $\tau_0$-WM is its ability to learn from heterogeneous interaction data, including real-robot teleoperation, UMI-style demonstrations, egocentric human videos, and simulator-oriented rollout trajectories. This unified training paradigm enables video prediction and action generation to be learned within a single framework while supporting deployment-time reasoning. Experiments on long-horizon, fine-grained manipulation tasks demonstrate strong policy performance across multiple robot embodiments, consistent gains from heterogeneous pre-training, and significant improvements from test-time computation.

Looking forward, several directions remain promising. First, many dexterous manipulation tasks require information beyond vision alone. Incorporating additional sensing modalities, particularly tactile feedback, may enable more reliable modeling of contact-rich interactions such as insertion, fastening, and deformable object manipulation. Second, while the proposed test-time computation strategy already improves execution performance, developing more reliable deployment-time reasoning mechanisms remains an important challenge. Better uncertainty estimation, longer-horizon evaluation, and more effective search strategies may further improve action selection in difficult states. Finally, extending predictive modeling to longer temporal horizons and more complex manipulation scenarios may enable richer future imagination and stronger decision-making capabilities.

Overall, we believe predictive robot learning provides a promising path toward more capable and reliable robot foundation models, and $\tau_0$-WM offers an practical step in this direction.




\section{Acknowledgement}

We would like to thank Jinyu Zhang, Sen Wang, Youlun Peng, Xinlin Ren, Mingjie Pan, Jianheng Song, Siyuan Feng, Zhongyuan Liu, Dong Li, Xiaowei Cai, Dafeng Wei, Han Jiang, Runkun Ju, Shaowei Li, Li Wang Buqing Nie, Kefeng Tang for their valuable contributions and support throughout this project. Their efforts in data collection, system development, experiment deployment, infrastructure maintenance, and engineering implementation were essential to the completion of this work.


\clearpage
\appendices
\appendix

\subsection{Training and Deploymeny Details}
\label{sec:appendix_training}

\subsubsection{Training Configuration}

$\tau_0$-WM is trained in two stages. The pre-training stage uses 27.3K hours of heterogeneous interaction data, including real-robot teleoperation, UMI-style interaction data, egocentric human videos, and rollout or failure trajectories. The post-training stage further adapts the model to downstream robotic manipulation tasks.

The pre-training and post-training stages use global batch sizes of 12,288 and 384, respectively. Both stages use the AdamW optimizer with a learning rate of $5\times10^{-5}$. Unless otherwise specified, all experiments use the same optimization hyperparameters across embodiments and tasks.

\subsubsection{Deployment Details}
All real-robot experiments are performed under language-conditioned multi-view observations. Unless otherwise stated, actions are executed in a receding-horizon closed-loop manner using fixed-length action chunks of 30.

Real-robot inference is deployed on a single RTX 5090 GPU. Under the standard deployment configuration, the end-to-end action generation latency is approximately 220\,ms per query. By caching reusable text representations, the latency can be reduced to approximately 180\,ms without changing model outputs.

\subsection{Inference Acceleration}
\label{sec:appendix_acceleration}

To improve deployment efficiency, we employ several implementation-level optimizations.

\subsubsection{Cross-Attention KV Cache}

During action-only inference, the video branch provides conditioning features for the action branch through cross-attention. Since the visual context remains unchanged throughout the denoising process, the corresponding key and value tensors are computed once and reused across all sampling steps.

Let $x^{(l)}$ denote the video feature at transformer layer $l$. We cache

\begin{equation}
K_v^{(l)} = W_K x^{(l)},
\qquad
V_v^{(l)} = W_V x^{(l)},
\end{equation}
and reuse them throughout inference, eliminating redundant projection operations.

\subsubsection{Fused QKV Projection}

We fuse query, key, and value projections into a single matrix multiplication,

\begin{equation}
[Q,K,V] = W_{QKV}x,
\end{equation}
which reduces kernel launch overhead and improves memory throughput compared with three independent projection layers.

\subsubsection{Simplified Rotary Position Embedding}

Action tokens form a one-dimensional temporal sequence. We therefore precompute one-dimensional rotary position embeddings and reuse them throughout deployment, avoiding repeated frequency construction and reducing positional encoding overhead.

\subsubsection{Torch Compile Optimization}

We additionally employ \texttt{torch.compile}~\cite{ansel2024pytorch2} with per-block compilation. Combined with the optimizations above, deployment latency can be further reduced from approximately 180\,ms to 140\,ms. Although \texttt{torch.compile} generally preserves model functionality, compiler-level graph transformations and kernel fusion may introduce small numerical differences compared with eager execution. For diffusion-based models, such differences can occasionally propagate through the sampling process and lead to slightly different outputs. Therefore, unless otherwise specified, all results reported in the main paper are obtained without \texttt{torch.compile} to ensure consistency and reproducibility across experiments.

\clearpage

\bibliographystyle{plainnat}
\bibliography{references}

\end{document}